\pdfoutput=1

\documentclass[11pt,a4paper]{article}
\usepackage[affil-sl,noblocks]{authblk}
\usepackage[hyperref]{acl2023} 
\usepackage{times}
\usepackage{latexsym}
\usepackage{mdframed}
\usepackage{graphicx}
\usepackage{tikzsymbols}
\usepackage{microtype}
\usepackage{amsmath}
\usepackage{csquotes}
\usepackage{booktabs}
\usepackage{tabularx}

\usepackage{natbib}

\usepackage[absolute,showboxes,overlay]{textpos}
\TPMargin{3pt}

\usepackage{soul}

\colorlet{cue}{violet!50}
\colorlet{source}{YellowOrange!50}
\colorlet{message}{Aquamarine!50}
\colorlet{sample}{gray!20}

\newcommand{\cue}[1]{\sethlcolor{cue}\hl{\emph{#1}}$_{\text{\textcolor{violet}{(Cue)}}}$}
\newcommand{\source}[1]{\sethlcolor{source}\hl{#1}$_{\text{\textcolor{YellowOrange}{(Source)}}}$}
\newcommand{\msg}[1]{\sethlcolor{message}\hl{#1}$_{\text{\textcolor{Aquamarine}{(Message)}}}$}
\newcommand{\sample}[1]{\sethlcolor{sample}\hl{#1}}

\title{Speaker Attribution in German Parliamentary Debates with QLoRA-adapted Large Language Models}

\author[1]{Tobias Bornheim}
\author[1,2,3]{Niklas Grieger}
\author[1]{Patrick Gustav Blaneck}
\author[1,3,*]{Stephan Bialonski}
\affil[1]{Department of Medical Engineering and Technomathematics\authorcr
FH Aachen University of Applied Sciences, Jülich, Germany\authorcr}
\affil[2]{Department of Information and Computing Sciences\authorcr
Utrecht University, Utrecht, The Netherlands\authorcr}
\affil[3]{Institute for Data-Driven Technologies\authorcr
FH Aachen University of Applied Sciences, Jülich, Germany}
\affil[*]{\textit{bialonski@fh-aachen.de}}

\date{}

\begin{document}
\maketitle
\begin{abstract}
  The growing body of political texts opens up new opportunities for rich insights into political dynamics and ideologies but also increases the workload for manual analysis.
  Automated speaker attribution, which detects who said what to whom in a speech event and is closely related to semantic role labeling, is an important processing step for computational text analysis.
  We study the potential of the large language model family Llama~2 to automate speaker attribution in German parliamentary debates from 2017--2021.
  We fine-tune Llama~2 with QLoRA, an efficient training strategy, and  observe our approach to achieve competitive performance in the Germ\-Eval 2023 Shared Task On Speaker Attribution in German News Articles and Parliamentary Debates.
  Our results shed light on the capabilities of large language models in automating speaker attribution, revealing a promising avenue for computational analysis of political discourse and the development of semantic role labeling systems.
\end{abstract}

\begin{textblock*}{17cm}[-0.1,0](0cm,27.8cm)
  \centering
  \small
  This work was published in J. Lang. Technol. Comput. Linguistics, 2024, available online at \url{https://jlcl.org/article/view/244}. 
  Please cite as: Bornheim, T., Grieger, N., Blaneck, P. G., \& Bialonski, S. (2024). Speaker Attribution in {G}erman Parliamentary debates with {QLoRA}-adapted Large Language Models. Journal for Language Technology and Computational Linguistics, 37, 1-13.
\end{textblock*}

\section{Introduction}

Language is central to the study of politics, as it forms the basis for political speech and debates \cite{Grimmer2013}. These textual sources offer rich insights into political dynamics and ideologies, yet the analysis of even moderately sized collections has been impeded by prohibitive costs. Recent innovations from natural language processing (NLP) have the potential to significantly reduce the financial burden of scrutinizing extensive text corpora \cite{Glavas2019,Abercrombie2020}. This development coincides with the availability of a growing body of political texts, including German Parliamentary data \cite{Barbaresi2018,Blatte2018,Walter2021,Rauh2020,Abrami2022,GermEval2023}, thus opening new avenues for political research.

Political texts are usually unstructured, presenting challenges for automated analyses. An approach towards this challenge is automated speaker attribution \cite{GermEval2023}, which detects who said what to whom in a speech event. This process involves detecting cue words that initiate a speech event and discerning the different roles (e.g., source, message, and addressee) associated with each event. This task is closely related to semantic role labeling (SRL) that delineates the specific semantic relationships among a predicate and its corresponding arguments, such as \enquote{who} did \enquote{what} to \enquote{whom}, \enquote{where}, \enquote{when}, and \enquote{why} \cite{Gildea2002,Marquez2008}. Semantic role labeling is considered a key component for natural language understanding and has been demonstrated to enhance systems for various applications including question answering, machine translation, and video understanding \cite{Navigli2022}.

Early approaches to SRL relied on syntactic features \cite{Navigli2022,Larionov2019}. More recently, the field has seen a significant transition from such engineered features to features learned in an end-to-end fashion by models that operate on raw-level input or tokens \cite{Collobert2011}. However, such end-to-end models necessitate large annotated training sets, available for English but scarce for low-resource languages. This problem can be mitigated by pretraining on unannotated data. Indeed, the emergence of pretrained large language models (LLMs) inspired by the transformer architecture \cite{Vaswani2017} led to new state-of-the-art results across various NLP tasks. Among these, encoder-only models like BERT were demonstrated to improve existing SRL benchmarks \cite{ShiP2019}. More recently, the advent of decoder-only models, such as GPT \cite{Radford2018} and larger models like GPT-4 \cite{GPT4-2023}, Claude~2 \cite{Bai2022}, and Llama~2 \cite{Touvron2023b}, has further propelled the field. These models, with their ability to comprehend and execute instructions in natural language for a wide array of tasks, hold potential for SRL and automated speaker attribution that is, to the best of our knowledge, largely unexplored.

In this contribution, we study the potential of Llama~2 70B, a model from a recently introduced family of large language models, to automatically detect speech events and attribute speakers in German parliamentary debates. We instruct and fine-tune Llama~2 to extract cues and roles using QLoRA \cite{Dettmers2023}, a parameter- and computationally efficient training strategy. Our approach achieves competitive performance (quantified by F1 scores for cues and roles) on the SpkAtt-2023 dataset of the \emph{GermEval 2023 Shared Task on Speaker Attribution in German News Articles and Parliamentary Debates} \cite{GermEval2023}. The implementation details of our experiments (Team \enquote{CPAa}) are available online\footnote{\url{https://github.com/dslaborg/germeval2023}}.

\section{Data and tasks}
\label{sec:data_and_tasks}

The dataset of the \emph{GermEval 2023 Shared Task on Speaker Attribution in German News Articles and Parliamentary Debates} consisted of 267 speeches from the German Bundestag \cite{GermEval2023}.
This dataset included speeches from all seven parliamentary groups (including independent members of parliament as a separate group) of the 19th legislative period of the German Bundestag (see Table \ref{tab:data_speeches_units_per_group} for details).
To facilitate analysis, each speech was automatically separated into sentence-like structures using spaCy, hereafter referred to as \emph{samples} (units of analysis).
Each sample was then further split into \emph{elements}, i.e., words and punctuation marks.

\begin{table}
    \centering
    \begin{tabularx}{\linewidth}{Xrr}
        \toprule
        \emph{Parliamentary group} & \emph{Speeches} & \emph{Samples} \\
        \midrule
        CDU/CSU                    & 77              & 4305           \\
        SPD                        & 57              & 2887           \\
        AfD                        & 39              & 1827           \\
        FDP                        & 34              & 1435           \\
        DIE LINKE                  & 29              & 1356           \\
        B'90 / DIE GRÜNEN          & 27              & 1152           \\
        independent                & 4               & 125            \\
        \midrule
        \emph{Total}               & 267             & 13087          \\
        \bottomrule
    \end{tabularx}
    \caption{Number of speeches and samples per parliamentary group in the combined \emph{Train}, \emph{Dev}, and \emph{Eval} datasets.}
    \label{tab:data_speeches_units_per_group}
\end{table}

Human annotators followed annotation guidelines \footnote{\url{https://github.com/umanlp/SpkAtt-2023/blob/master/doc/Guidelines_SpeakerAttribution_in_Parliamentary_Debates-SpkAtt-2023_Task1.pdf}} to assign none, one, or multiple annotations to each sample.
These annotations consisted of \emph{cue words} that invoke speech events and roles (\emph{Addr}, \emph{Evidence}, \emph{Medium}, \emph{Message}, \emph{Source}, \emph{Topic}, \emph{PTC}) associated with that event.
While the cue is mandatory for each annotation, roles are context-dependent and may be absent.
Figure \ref{fig:example_annotation} shows example annotations.

\begin{figure}[t]
    \centering
    \begin{mdframed}[
            leftmargin=0pt,
            rightmargin=0pt,
            skipabove=0pt,
            skipbelow=0pt,
        ]
        \small
        \emph{Annotation 1}
        \\
        Von der AfD wollen wir hier lieber nicht reden; \ddag\, %
        denn \source{wir} \cue{wissen}: \msg{Neben ihren rassistischen Positionen \ddag\, %
            haben die Rechtsradikalen nicht nur Klimawandelleugnung im Angebot, sie haben auch die rechtspopulistischen Positionen eines Donald Trump gepachtet}.
        \\

        \emph{Annotation 2}
        \\
        Von der AfD wollen wir hier lieber nicht reden; \ddag\, %
        denn wir wissen: Neben ihren rassistischen \cue{Positionen} \ddag\, %
        haben die Rechtsradikalen nicht nur Klimawandelleugnung im Angebot, sie haben auch die rechtspopulistischen Positionen eines Donald Trump gepachtet.
        \\

        \emph{Annotation 3}
        \\
        Von der AfD wollen wir hier lieber nicht reden; \ddag\, %
        denn \source{wir} wissen: Neben ihren rassistischen Positionen \ddag\, %
        haben die Rechtsradikalen nicht nur Klimawandelleugnung im Angebot, sie haben auch die rechtspopulistischen \cue{Positionen} \msg{eines Donald Trump gepachtet}.
    \end{mdframed}
    \caption{ %
        Sentence from the \emph{Train} dataset with three annotations. %
        The sentence was split into three samples by spaCy (splitting points are indicated by \ddag). %
        This segmentation also occurs at not-punctuated positions, as seen in the example sentence (\enquote{\ldots rassistischen Positionen \ddag\, haben die Rechtsradikalen \ldots}). %
        This behavior is due to the data provided by \enquote{Open Bundestag}, where comments from other members of parliament during an otherwise coherent paragraph force this unintuitive segmentation into two separate paragraphs \protect\cite{GermEval2023}. %
        As seen in \emph{Annotation 2}, there can be annotations consisting of only cue word(s). %
        \emph{Annotation 1} and \emph{Annotation 3} show that annotated roles can span multiple samples. %
    }
    \label{fig:example_annotation}
\end{figure}

The Shared Task consisted of two subtasks: \emph{Full Annotation} (\emph{Subtask 1}) and \emph{Role Detection} (\emph{Subtask 2}) \cite{GermEval2023}.
In the \emph{Full Annotation} subtask, the goal was to predict all cues and roles for each sample.
In the \emph{Role Detection} subtask, the gold cues were given, and the goal was to predict only the roles for each sample.

The dataset was provided as five sets, namely \emph{Trial}, \emph{Train}, \emph{Dev}, and two \emph{Eval} sets (see Table \ref{tab:data_speeches_units_per_split}).
We omitted the \emph{Trial} set in our experiments, since it was included in the \emph{Train} set.
For training and tuning the final models, we used the \emph{Train} and \emph{Dev} sets.
The two \emph{Eval} sets were used by the GermEval 2023 organizers to compute the final scores for Subtask 1 (\emph{Eval} set 1) and Subtask 2 (\emph{Eval} set 2).
While the two \emph{Eval} sets contained the same samples, the organizers provided gold cues with \emph{Eval} set 2.

\begin{table}
    \centering
    \begin{tabularx}{\linewidth}{Xrrr}
        \toprule
        \emph{Split} & \emph{Speeches} & \emph{Samples} & \emph{Annotations} \\
        \midrule
        Dev          & 18              & 927            & 515                \\
        Train        & 177             & 9093           & 5399               \\
        Eval         & 72              & 3067           & 1792               \\
        \midrule
        \emph{Total} & 267             & 13087          & 7706               \\
        \bottomrule
    \end{tabularx}
    \caption{Number of speeches, samples (units of analysis), and annotations for each dataset.
        The \emph{Trial} dataset is completely contained within the \emph{Train} dataset and is therefore not shown.
        The \emph{Eval} dataset here refers to the test sets of both \emph{Subtask 1} and \emph{Subtask 2}, since they only differ in the provided annotations.
    }
    \label{tab:data_speeches_units_per_split}
\end{table}

\section{Methods}
\label{sec:methods}

\subsection{Models}
\label{ssec:models}

We used the Llama~2 model family \cite{Touvron2023b}, a set of large language models pretrained on a corpus of two trillion tokens with a context length of 4096 tokens.
The Llama~2 model family includes both pretrained models and fine-tuned versions optimized for conversational tasks.
Since our approach did not require the conversational capabilities of the fine-tuned models, we chose to use the base pretrained versions of Llama~2 in our experiments.
These base models were trained without a specific prompt format and are therefore not biased toward any particular prompt strategy, allowing us to freely choose our own prompt format.

While the Llama~2 model family contains models of various sizes, we chose to fine-tune the largest available model with 70 billion parameters (Llama~2 70B).
The weights of this model can be obtained upon request using the official GitHub repository\footnote{\url{https://github.com/facebookresearch/llama}}.
Once downloaded, we followed the provided instructions\footnote{\url{https://github.com/facebookresearch/llama-recipes}} to convert the model to the HuggingFace Transformers format \cite{Wolf2020}.
This conversion allowed us to load the model using the HuggingFace Transformers library, which facilitated the fine-tuning and inference steps.

\subsection{Preprocessing}
\label{sec:preprocessing}

For effective training (see section \ref{sec:training}) and inference (see section \ref{sec:inference}) we preprocessed each sample.
We parsed each annotation into its respective lists of elements.
Next, we joined all elements of a sample with space characters in between to get each sample's \emph{text}.
Since roles can be contained in samples different from the one containing the cue, we concatenated the sample with the next two samples of the same speech, if possible.

During our experiments, we noticed that our models ignored their instructions and generated random text if the text of a given sample ended with a colon.
To counteract this behavior, we replaced this trailing colon with a period.

\begin{figure}[t]
    \centering
    \begin{mdframed}
        \small
        \emph{Input:}\\
        User: A cue is the lexical items in a sentence that indicate that speech, writing, or thought is being reproduced.\\
        I want you to extract all cues in the text below.\\
        If you find multiple words for one cue, you output them separated by commas.\\
        If no cue can be found in the given text, you output the string \#UNK\# as cue.\\
        Now extract all cues from the following sentence.\\
        Use the prefix ``Cues: ''.\\
        Sentence: \sample{denn wir wissen: Neben ihren rassistischen Positionen}\\
        Assistant:\\
        \\
        \emph{Output:}\\
        Cues: \sample{[wissen], [Positionen]}\textless/s\textgreater
    \end{mdframed}
    \caption{Example cue prompt and desired model response for the sample ``denn wir wissen: Neben ihren rassistischen Positionen'' with the cues ``wissen'' and ``Positionen''.
        Shaded in gray are the parts of the prompt and response that are sample dependent.
        The prompt is used as the \emph{Input} sequence for training and inference, while the \emph{Output} sequence contains the desired response with the cues.
        The end-of-sentence token ``\textless/s\textgreater'' is used to indicate the end of the \emph{Output} sequence.
    }
    \label{fig:cue_prompt}
\end{figure}

\begin{figure}[t]
    \centering
    \begin{mdframed}
        \small
        \emph{Input:}\\
        User: Now I give you again the sentence only in addition with the two following sentences, because the roles can be partially contained in the following sentences.\\
        Text: \sample{denn wir wissen : Neben ihren rassistischen Positionen} \ddag\, \sample{haben die Rechtsradikalen nicht nur Klimawandelleugnung im Angebot , sie haben auch die rechtspopulistischen Positionen eines Donald Trump gepachtet .} \ddag\, \sample{Als Linke \"ubernehmen wir Verantwortung .} \\
        Now find all roles in the sentence associated with the cue `\sample{wissen}' you found in the beginning sentence.\\
        Assistant:\\
        \\
        \emph{Output:}\\
        cue: \sample{wissen}\\
        ptc: \sample{\#UNK\#}\\
        evidence: \sample{\#UNK\#}\\
        medium: \sample{\#UNK\#}\\
        topic: \sample{\#UNK\#}\\
        addr: \sample{\#UNK\#}\\
        message: \sample{Neben, ihren, rassistischen, Positionen, haben, die, Rechtsradikalen, nicht, nur, Klimawandelleugnung, im, Angebot, ,, sie, haben, auch, die, rechtspopulistischen, Positionen, eines, Donald, Trump, gepachtet}\\
        source: \sample{wir}\textless/s\textgreater
    \end{mdframed}
    \caption{Example role prompt and desired model response for the sample ``denn wir wissen: Neben ihren rassistischen Positionen'' with the cue ``wissen''.
        Since roles can be contained in samples different from the one containing the cue, we concatenated the sample with the next two samples of the same speech (transitions between samples are indicated by \ddag).
        Shaded in gray are the parts of the prompt and response that are sample dependent.
        Similar to the cue prompt, the role prompt is used as the \emph{Input} sequence for training and inference, while the \emph{Output} sequence contains the desired response.
        We append the end-of-sentence token ``\textless/s\textgreater'' to the \emph{Output}.
    }
    \label{fig:role_prompt}
\end{figure}

We designed prompts for cue prompting (see Figure \ref{fig:cue_prompt}) and role prompting (see Figure \ref{fig:role_prompt}).
We wrote the instructions in our prompt templates in English, because it was observed that the performance of multilingual models such as Llama~2 is improved when English prompts are used \cite{Fu2022,Huang2023}.
Also, since a sample may not contain a cue, or a role may be missing, we used ``\#UNK\#'' to mark such cases.

\subsection{Training}
\label{sec:training}

For our final submission, we fine-tuned two Llama~2 70B models to identify cues and roles, respectively, using QLoRA (Quantized Low-Rank Adaptation) \cite{Dettmers2023}.
QLoRA is a highly efficient fine-tuning technique for large language models that achieves similar performance to full fine-tuning while using only a fraction of the memory.
This memory reduction is achieved by quantizing the model weights of an LLM to four bits and adding Low Rank Adapters (LoRA layers) to all linear transformer blocks of the model.
During fine-tuning, only these LoRA layers are trained and the rest of the pretrained model weights remain unaltered.
By employing this strategy, QLoRA achieves a significant reduction in memory usage during fine-tuning, while still allowing the model to adapt to downstream tasks through the trainable LoRA layers.

As described in Section \ref{sec:preprocessing}, we parsed the training samples into cue prompts (see Figure \ref{fig:cue_prompt}) that served as input to the cue model and role prompts (see Figure \ref{fig:role_prompt}) that served as input to the role model.
Utilizing these input prompts, the respective models were trained to predict the desired assistant responses (defined as \emph{Output} in Figures \ref{fig:cue_prompt} and \ref{fig:role_prompt}).
This approach is consistent with previous research that has shown improved performance when fine-tuning only on the target response of an instruction set, rather than both the instructions and the desired response \cite{Dettmers2023}.
By treating the input and output separately, we can process the two sequences with different maximum sequence lengths.
Specifically, for the model used to identify cues, we set the maximum length of the input to 256 tokens (with seven samples of the training data truncated) and the maximum length of the output to 64 tokens (no samples truncated).
For the model used to identify roles, we truncated the input to 640 tokens (with six samples of the training data truncated) and the output to 256 tokens (with one sample truncated).

Except for the maximum number of tokens in the input and output sequences, we largely followed the training strategy proposed in \citet*{Dettmers2023}.
Although their specific experiments did not involve a Llama~2 70B model, they successfully fine-tuned a similarly sized LLaMA model (predecessor to Llama~2) with 65 billion parameters \cite{Touvron2023a}.
We adopted most parameters from this 65B model fine-tuning, such as a constant learning rate of $\eta=0.0001$ with linear warmup over the first 3\% of training steps and a dropout of 0.05 for the LoRA layers.
The main hyperparameter we adjusted was the number of training steps to prevent overfitting.
For the cues model, we trained for 2000 steps with a batch size of 16 and no gradient accumulation.
For the roles model, we used 2500 steps with a batch size of eight and gradient accumulation over two steps, i.e., an effective batch size of 16.

Fine-tuning was carried out on a DGX A100 server, with a total training time of about seven hours for the cues model and 17 hours for the roles model.
To optimize memory usage, we experimented with reducing the batch size to one while increasing the gradient accumulation steps to 16 (i.e., maintaining the same effective batch size).
With these parameters, both models were able to operate within a GPU memory limit of less than 60 GB.

\subsection{Inference}
\label{sec:inference}

Prompting our fine-tuned models was a two-step process.
In the first step, we prompted our cue model for all cues in a sample using our prompt template for cues (see Figure \ref{fig:cue_prompt}).
We postprocessed the output of the model (see section \ref{sec:postprocessing}) into a list of cues.
In the second step, for each cue, we prompted for the roles with our role model.
To do this, we prepended the complete cue prompt and its output to the role prompt template before querying the model (see Figure \ref{fig:role_prompt}).

To ensure reproducibility of results, we configured our models to generate output deterministically.
For a given input sequence, large language models obtain a probability distribution over all possible tokens.
We chose to always select the token with the highest assigned probability as the next output token, thereby fixing the output for a given input sequence.

\subsection{Postprocessing and evaluation metrics}
\label{sec:postprocessing}

Several postprocessing steps were necessary to evaluate the models' output in a structured way.

\paragraph{Enforcing the output format.} If the models' output did not follow our strict output format (see Figures \ref{fig:cue_prompt} and \ref{fig:role_prompt}), we mapped the output to the marker \#UNK\# (unknown).

\paragraph{Preventing overlapping cues.} If our cue model detected multiple but overlapping cues, we combined them into a single cue.

\paragraph{Ignoring made-up words.} If the output of the model contained words for cues or roles that were not in the given sample, and no other word with a Levenshtein distance of 1 was found in the sample, we ignored those words. Then, if the output was empty, we mapped the output to the marker \#UNK\# (unknown).

\paragraph{Resolving ambiguities.} A word may occur more than once in a sample.
When a model outputs such a word as a cue or a role, it is unclear to which occurrence of the word in the sample it should be attributed.
To resolve this ambiguity, for each occurrence of the word, we counted how many elements around that word (in the range of two elements to the left and right) were part of the cue or role, and chose the occurrence with the highest count.

\paragraph{Including surrounded punctuation.} Roles often contained punctuation marks such as colons or commas.
We observed that our models ignored these punctuation marks most of the time.
If a punctuation mark was surrounded by words that were selected for this role, we added that punctuation mark to the role as well.

\paragraph{Evaluating metrics.}
To evaluate the performance of our models, we used the proportional F1 score as proposed for opinion role labeling \cite{Johansson2010}.
This score is defined as the harmonic mean of the proportional precision and recall.
Proportional precision quantifies the proportion of overlap between a predicted cue (role) and an overlapping true cue (role).
Proportional recall quantifies the proportion of overlap between a true cue (role) and an overlapping predicted cue (role; see \citet*{GermEval2023} for further details on how the proportional F1 score is calculated).

\section{Results}

\begin{table}
  \centering
  \begin{tabularx}{\linewidth}{Xrrr}
    \toprule
                  & \emph{Precision} & \emph{Recall} & \emph{F1} \\
    \midrule
    \multicolumn{4}{l}{\emph{Subtask 1}}                         \\
    Cues          & 0.889            & 0.889         & 0.889     \\
    Roles         & 0.787            & 0.822         & 0.804     \\
    Cues \& Roles & 0.798            & 0.829         & 0.813     \\
    \midrule
    \multicolumn{4}{l}{\emph{Subtask 2}}                         \\
    Roles         & 0.910            & 0.873         & 0.891     \\
    \bottomrule
  \end{tabularx}
  \caption{Proportional precision, recall, and F1 scores obtained for predicting cues and roles on the \emph{Eval} dataset.
    The joint scores for predicting both cues and roles (Subtask 1 of GermEval 2023 Shared Task 1) are shown in the third row.
    The last row shows the results obtained for predicting roles on the \emph{Eval} dataset when the true cues were given (Subtask 2).
  }
  \label{tab:f1scores}
\end{table}

We used the same fine-tuned Llama~2 70B models for both Subtask 1 and Subtask 2 of GermEval 2023 Shared Task 1 -- a cues model to identify cues in a given sentence and a roles model to predict the roles associated with the identified cues.
While the cues model was used exclusively in Subtask 1, as the cues were provided in Subtask 2, the roles model was used in both subtasks.
It leveraged either the predicted cues from Subtask 1 or the gold cues from Subtask 2 to predict the roles associated with each cue, as described in section \ref{sec:inference}.
By using the same fine-tuned roles model for both subtasks, we were able to analyze the impact of using gold cues versus predicted cues on role identification performance.

Table \ref{tab:f1scores} shows the final results of our submissions on the \emph{Eval} dataset, as reported by the organizers of the GermEval 2023 Shared Task.
For Subtask~1, the fine-tuned cues model achieved an F1 score of 0.889 for predicting cues.
Using the predicted cues from this model, the fine-tuned roles model achieved an F1 score of 0.804 for predicting roles.
Combining both predictions, our models achieved an overall F1 score of 0.813 for predicting cues and roles in Subtask 1.
In Subtask 2, where gold cues were provided, the same roles model used in Subtask 1 achieved a higher F1 score of 0.891 for predicting roles.
Interestingly, the improvement of the roles model using gold cues was greater in precision, which increased from 0.787 to 0.910, than in recall, which increased from 0.822 to 0.873.
This increase in precision suggests that the cues model in Subtask 1 overpredicted sentences as containing cues when they actually had no cues, resulting in too many false positive role predictions.

In summary, our results demonstrate that our fine-tuned models are effective at reliably predicting cues and roles.
Additionally, the results highlight the importance of accurate cue prediction, as errors of the cues model propagate to the roles model, reducing its performance.

\section{Conclusion}

We demonstrated that fine-tuned Llama~2 language models can successfully predict cues and roles in German parliamentary debates, achieving competitive performance on the GermEval2023 Shared Task without relying on traditional linguistic features. These results highlight the feasibility of automated speaker attribution by fine-tuning models on prompt templates that task them with identifying cues and roles. The similarity between automated speaker attribution and semantic role labeling suggests that this strategy may pave the way for new state-of-the-art results in various semantic role labeling tasks.

\section*{Limitations}

We did not study risks that may or may not arise when our fine-tuned large language models are used for other application scenarios than ours. In our approach, users can neither manipulate the prompts nor read the generated texts produced by our models. Instead, the generated outputs are processed and mapped back to the words from the parliamentary speeches used as input. Therefore, we consider the risks associated with our approach to be limited. We recommend security testing if our trained models are to be used in other scenarios.

\section*{Acknowledgements}
We are grateful to M. Reißel and V. Sander for providing us with computing resources.

\appendix


\begin{thebibliography}{26}
  \expandafter\ifx\csname natexlab\endcsname\relax\def\natexlab#1{#1}\fi

  \bibitem[{Abercrombie and Batista-Navarro(2020)}]{Abercrombie2020}
  Gavin Abercrombie and Riza Batista-Navarro. 2020.
  \newblock \href {https://doi.org/10.1007/s42001-019-00060-w} {Sentiment and
  position-taking analysis of parliamentary debates: {A} systematic literature
  review}.
  \newblock \emph{Journal of Computational Social Science}, 3(1):245--270.

  \bibitem[{Abrami et~al.(2022)Abrami, Bagci, Hammerla, and Mehler}]{Abrami2022}
  Giuseppe Abrami, Mevl{\"{u}}t Bagci, Leon Hammerla, and Alexander Mehler. 2022.
  \newblock \href {https://aclanthology.org/2022.lrec-1.202} {German
  parliamentary corpus ({G}er{P}ar{C}or)}.
  \newblock In \emph{Proceedings of the Thirteenth Language Resources and
    Evaluation Conference, {LREC} 2022, Marseille, France, 20-25 June 2022},
  pages 1900--1906. European Language Resources Association.

  \bibitem[{Bai et~al.(2022)Bai, Kadavath, Kundu, Askell, Kernion, Jones, Chen,
        Goldie, Mirhoseini, McKinnon, Chen, Olsson, Olah, Hernandez, Drain, Ganguli,
        Li, Tran{-}Johnson, Perez, Kerr, Mueller, Ladish, Landau, Ndousse, Lukosiute,
        Lovitt, Sellitto, Elhage, Schiefer, Mercado, DasSarma, Lasenby, Larson,
        Ringer, Johnston, Kravec, Showk, Fort, Lanham, Telleen{-}Lawton, Conerly,
        Henighan, Hume, Bowman, Hatfield{-}Dodds, Mann, Amodei, Joseph, McCandlish,
        Brown, and Kaplan}]{Bai2022}
  Yuntao Bai, Saurav Kadavath, Sandipan Kundu, Amanda Askell, Jackson Kernion,
  Andy Jones, Anna Chen, Anna Goldie, Azalia Mirhoseini, Cameron McKinnon,
  Carol Chen, Catherine Olsson, Christopher Olah, Danny Hernandez, Dawn Drain,
  Deep Ganguli, Dustin Li, Eli Tran{-}Johnson, Ethan Perez, Jamie Kerr, Jared
  Mueller, Jeffrey Ladish, Joshua Landau, Kamal Ndousse, Kamile Lukosiute,
  Liane Lovitt, Michael Sellitto, Nelson Elhage, Nicholas Schiefer,
  Noem{\'{\i}} Mercado, Nova DasSarma, Robert Lasenby, Robin Larson, Sam
  Ringer, Scott Johnston, Shauna Kravec, Sheer~El Showk, Stanislav Fort, Tamera
  Lanham, Timothy Telleen{-}Lawton, Tom Conerly, Tom Henighan, Tristan Hume,
  Samuel~R. Bowman, Zac Hatfield{-}Dodds, Ben Mann, Dario Amodei, Nicholas
  Joseph, Sam McCandlish, Tom Brown, and Jared Kaplan. 2022.
  \newblock \href {https://doi.org/10.48550/arXiv.2212.08073} {Constitutional
      {AI:} {H}armlessness from {AI} feedback}.
  \newblock \emph{CoRR}, abs/2212.08073.

  \bibitem[{Barbaresi(2018)}]{Barbaresi2018}
  Adrien Barbaresi. 2018.
  \newblock \href {https://aclanthology.org/L18-1127} {A corpus of {G}erman
    political speeches from the 21st century}.
  \newblock In \emph{Proceedings of the Eleventh International Conference on
    Language Resources and Evaluation ({LREC} 2018)}, Miyazaki, Japan. European
  Language Resources Association (ELRA).

  \bibitem[{Bl{\"{a}}tte and Blessing(2018)}]{Blatte2018}
  Andreas Bl{\"{a}}tte and Andr{\'{e}} Blessing. 2018.
  \newblock \href
  {http://www.lrec-conf.org/proceedings/lrec2018/summaries/1024.html} {The
    {G}erma{P}arl corpus of parliamentary protocols}.
  \newblock In \emph{Proceedings of the Eleventh International Conference on
    Language Resources and Evaluation, {LREC} 2018, Miyazaki, Japan, May 7-12,
    2018}. European Language Resources Association {(ELRA)}.

  \bibitem[{Collobert et~al.(2011)Collobert, Weston, Bottou, Karlen, Kavukcuoglu,
        and Kuksa}]{Collobert2011}
  Ronan Collobert, Jason Weston, L{\'{e}}on Bottou, Michael Karlen, Koray
  Kavukcuoglu, and Pavel~P. Kuksa. 2011.
  \newblock \href {https://doi.org/10.5555/1953048.2078186} {Natural language
    processing (almost) from scratch}.
  \newblock \emph{J. Mach. Learn. Res.}, 12:2493--2537.

  \bibitem[{Dettmers et~al.(2023)Dettmers, Pagnoni, Holtzman, and
        Zettlemoyer}]{Dettmers2023}
  Tim Dettmers, Artidoro Pagnoni, Ari Holtzman, and Luke Zettlemoyer. 2023.
  \newblock \href {https://doi.org/10.48550/arXiv.2305.14314} {{QLoRA}:
  {E}fficient finetuning of quantized {LLMs}}.
  \newblock \emph{CoRR}, abs/2305.14314.

  \bibitem[{Fu et~al.(2022)Fu, Ng, and Liu}]{Fu2022}
  Jinlan Fu, See-Kiong Ng, and Pengfei Liu. 2022.
  \newblock \href {https://doi.org/10.18653/v1/2022.emnlp-main.674} {Polyglot
    {P}rompt: {M}ultilingual multitask prompt training}.
  \newblock In \emph{Proceedings of the 2022 Conference on Empirical Methods in
    Natural Language Processing}, pages 9919--9935, Abu Dhabi, United Arab
  Emirates. Association for Computational Linguistics.

  \bibitem[{Gildea and Jurafsky(2002)}]{Gildea2002}
  Daniel Gildea and Daniel Jurafsky. 2002.
  \newblock \href {https://doi.org/10.1162/089120102760275983} {Automatic
    labeling of semantic roles}.
  \newblock \emph{Comput. Linguistics}, 28(3):245--288.

  \bibitem[{Glava{\v{s}} et~al.(2019)Glava{\v{s}}, Nanni, and
        Ponzetto}]{Glavas2019}
  Goran Glava{\v{s}}, Federico Nanni, and Simone~Paolo Ponzetto. 2019.
  \newblock \href {https://doi.org/10.18653/v1/p19-4004} {Computational analysis
    of political texts: Bridging research efforts across communities}.
  \newblock In \emph{Proceedings of the 57th Annual Meeting of the Association
    for Computational Linguistics: Tutorial Abstracts}. Association for
  Computational Linguistics.

  \bibitem[{Grimmer and Stewart(2013)}]{Grimmer2013}
  Justin Grimmer and Brandon~M. Stewart. 2013.
  \newblock \href {https://doi.org/10.1093/pan/mps028} {Text as data: {T}he
  promise and pitfalls of automatic content analysis methods for political
  texts}.
  \newblock \emph{Political Analysis}, 21(3):267--297.

  \bibitem[{Huang et~al.(2023)Huang, Tang, Zhang, Zhao, Song, Xia, and
        Wei}]{Huang2023}
  Haoyang Huang, Tianyi Tang, Dongdong Zhang, Wayne~Xin Zhao, Ting Song, Yan Xia,
  and Furu Wei. 2023.
  \newblock \href {https://doi.org/10.48550/arXiv.2305.07004} {Not all languages
  are created equal in {LLMs}: {I}mproving multilingual capability by
  cross-lingual-thought prompting}.
  \newblock \emph{CoRR}, abs/2305.07004.

  \bibitem[{Johansson and Moschitti(2010)}]{Johansson2010}
  Richard Johansson and Alessandro Moschitti. 2010.
  \newblock \href {https://aclanthology.org/W10-2910} {Syntactic and semantic
    structure for opinion expression detection}.
  \newblock In \emph{Proceedings of the Fourteenth Conference on Computational
    Natural Language Learning}, pages 67--76, Uppsala, Sweden. Association for
  Computational Linguistics.

  \bibitem[{Larionov et~al.(2019)Larionov, Shelmanov, Chistova, and
        Smirnov}]{Larionov2019}
  Daniil Larionov, Artem Shelmanov, Elena Chistova, and Ivan~V. Smirnov. 2019.
  \newblock \href {https://doi.org/10.26615/978-954-452-056-4\_073} {Semantic
    role labeling with pretrained language models for known and unknown
    predicates}.
  \newblock In \emph{Proceedings of the International Conference on Recent
    Advances in Natural Language Processing, {RANLP} 2019, Varna, Bulgaria,
    September 2-4, 2019}, pages 619--628. {INCOMA} Ltd.

  \bibitem[{M{\`{a}}rquez et~al.(2008)M{\`{a}}rquez, Carreras, Litkowski, and
  Stevenson}]{Marquez2008}
  Llu{\'{\i}}s M{\`{a}}rquez, Xavier Carreras, Kenneth~C. Litkowski, and Suzanne
  Stevenson. 2008.
  \newblock \href {https://doi.org/10.1162/coli.2008.34.2.145} {Semantic role
    labeling: An introduction to the special issue}.
  \newblock \emph{Comput. Linguistics}, 34(2):145--159.

  \bibitem[{Navigli et~al.(2022)Navigli, Barba, Conia, and
        Blloshmi}]{Navigli2022}
  Roberto Navigli, Edoardo Barba, Simone Conia, and Rexhina Blloshmi. 2022.
  \newblock \href {https://aclanthology.org/2022.aacl-tutorials.6} {A tour of
    explicit multilingual semantics: Word sense disambiguation, semantic role
    labeling and semantic parsing}.
  \newblock In \emph{Proceedings of the 2nd Conference of the Asia-Pacific
    Chapter of the Association for Computational Linguistics and the 12th
    International Joint Conference on Natural Language Processing: Tutorial
    Abstracts}, pages 35--43, Taipei. Association for Computational Linguistics.

  \bibitem[{OpenAI(2023)}]{GPT4-2023}
  OpenAI. 2023.
  \newblock \href {https://doi.org/10.48550/arXiv.2303.08774} {{GPT-4} technical
    report}.
  \newblock \emph{CoRR}, abs/2303.08774.

  \bibitem[{Radford and Narasimhan(2018)}]{Radford2018}
  Alec Radford and Karthik Narasimhan. 2018.
  \newblock \href
  {https://cdn.openai.com/research-covers/language-unsupervised/language_understanding_paper.pdf}
  {Improving language understanding by generative pre-training}.

  \bibitem[{Rauh and Schwalbach(2020)}]{Rauh2020}
  Christian Rauh and Jan Schwalbach. 2020.
  \newblock \href {https://doi.org/10.7910/DVN/L4OAKN} {{The ParlSpeech V2 data
        set: Full-text corpora of 6.3 million parliamentary speeches in the key
        legislative chambers of nine representative democracies}}.

  \bibitem[{Rehbein et~al.(2023)Rehbein, Petersen-Frey, Brunner, Ruppenhofer,
        Biemann, and Ponzetto}]{GermEval2023}
  Ines Rehbein, Fynn Petersen-Frey, Annelen Brunner, Josef Ruppenhofer, Chris
  Biemann, and Simone~Paolo Ponzetto. 2023.
  \newblock {Overview of the Germ\-Eval 2023 Shared Task on Speaker Attribution
    in Newswire and Parliamentary Debates}.
  \newblock In \emph{The Germ\-Eval 2023 Shared Task at KONVENS 2023},
  Ingolstadt, Germany.

  \bibitem[{Shi and Lin(2019)}]{ShiP2019}
  Peng Shi and Jimmy Lin. 2019.
  \newblock \href {http://arxiv.org/abs/1904.05255} {Simple {BERT} models for
    relation extraction and semantic role labeling}.
  \newblock \emph{CoRR}, abs/1904.05255.

  \bibitem[{Touvron et~al.(2023{\natexlab{a}})Touvron, Lavril, Izacard, Martinet,
  Lachaux, Lacroix, Rozi{\`{e}}re, Goyal, Hambro, Azhar, Rodriguez, Joulin,
  Grave, and Lample}]{Touvron2023a}
  Hugo Touvron, Thibaut Lavril, Gautier Izacard, Xavier Martinet, Marie{-}Anne
  Lachaux, Timoth{\'{e}}e Lacroix, Baptiste Rozi{\`{e}}re, Naman Goyal, Eric
  Hambro, Faisal Azhar, Aur{\'{e}}lien Rodriguez, Armand Joulin, Edouard Grave,
  and Guillaume Lample. 2023{\natexlab{a}}.
  \newblock \href {https://doi.org/10.48550/arXiv.2302.13971} {{LLaMA}: {O}pen
  and efficient foundation language models}.
  \newblock \emph{CoRR}, abs/2302.13971.

  \bibitem[{Touvron et~al.(2023{\natexlab{b}})Touvron, Martin, Stone, Albert,
        Almahairi, Babaei, Bashlykov, Batra, Bhargava, Bhosale, Bikel, Blecher,
        Canton{-}Ferrer, Chen, Cucurull, Esiobu, Fernandes, Fu, Fu, Fuller, Gao,
        Goswami, Goyal, Hartshorn, Hosseini, Hou, Inan, Kardas, Kerkez, Khabsa,
        Kloumann, Korenev, Koura, Lachaux, Lavril, Lee, Liskovich, Lu, Mao, Martinet,
        Mihaylov, Mishra, Molybog, Nie, Poulton, Reizenstein, Rungta, Saladi,
        Schelten, Silva, Smith, Subramanian, Tan, Tang, Taylor, Williams, Kuan, Xu,
        Yan, Zarov, Zhang, Fan, Kambadur, Narang, Rodriguez, Stojnic, Edunov, and
        Scialom}]{Touvron2023b}
  Hugo Touvron, Louis Martin, Kevin Stone, Peter Albert, Amjad Almahairi, Yasmine
  Babaei, Nikolay Bashlykov, Soumya Batra, Prajjwal Bhargava, Shruti Bhosale,
  Dan Bikel, Lukas Blecher, Cristian Canton{-}Ferrer, Moya Chen, Guillem
  Cucurull, David Esiobu, Jude Fernandes, Jeremy Fu, Wenyin Fu, Brian Fuller,
  Cynthia Gao, Vedanuj Goswami, Naman Goyal, Anthony Hartshorn, Saghar
  Hosseini, Rui Hou, Hakan Inan, Marcin Kardas, Viktor Kerkez, Madian Khabsa,
  Isabel Kloumann, Artem Korenev, Punit~Singh Koura, Marie{-}Anne Lachaux,
  Thibaut Lavril, Jenya Lee, Diana Liskovich, Yinghai Lu, Yuning Mao, Xavier
  Martinet, Todor Mihaylov, Pushkar Mishra, Igor Molybog, Yixin Nie, Andrew
  Poulton, Jeremy Reizenstein, Rashi Rungta, Kalyan Saladi, Alan Schelten, Ruan
  Silva, Eric~Michael Smith, Ranjan Subramanian, Xiaoqing~Ellen Tan, Binh Tang,
  Ross Taylor, Adina Williams, Jian~Xiang Kuan, Puxin Xu, Zheng Yan, Iliyan
  Zarov, Yuchen Zhang, Angela Fan, Melanie Kambadur, Sharan Narang,
  Aur{\'{e}}lien Rodriguez, Robert Stojnic, Sergey Edunov, and Thomas Scialom.
  2023{\natexlab{b}}.
  \newblock \href {https://doi.org/10.48550/arXiv.2307.09288} {Llama 2: Open
    foundation and fine-tuned chat models}.
  \newblock \emph{CoRR}, abs/2307.09288.

  \bibitem[{Vaswani et~al.(2017)Vaswani, Shazeer, Parmar, Uszkoreit, Jones,
        Gomez, Kaiser, and Polosukhin}]{Vaswani2017}
  Ashish Vaswani, Noam Shazeer, Niki Parmar, Jakob Uszkoreit, Llion Jones,
  Aidan~N Gomez, {\L}ukasz Kaiser, and Illia Polosukhin. 2017.
  \newblock \href {http://arxiv.org/abs/1706.03762} {Attention is all you need}.
  \newblock In \emph{Annual Conf. Neural Information Processing Systems 2017},
  pages 5998--6008, Long Beach, CA, USA.

  \bibitem[{Walter et~al.(2021)Walter, Kirschner, Eger, Glavas, Lauscher, and
        Ponzetto}]{Walter2021}
  Tobias Walter, Celina Kirschner, Steffen Eger, Goran Glavas, Anne Lauscher, and
  Simone~Paolo Ponzetto. 2021.
  \newblock \href {https://doi.org/10.1109/JCDL52503.2021.00017} {Diachronic
  analysis of {G}erman parliamentary proceedings: {I}deological shifts through
  the lens of political biases}.
  \newblock In \emph{{ACM/IEEE} Joint Conference on Digital Libraries, {JCDL}
    2021, Champaign, IL, USA, September 27-30, 2021}, pages 51--60. {IEEE}.

  \bibitem[{Wolf et~al.(2020)Wolf, Debut, Sanh, Chaumond, Delangue, Moi, Cistac,
        Rault, Louf, Funtowicz, Davison, Shleifer, von Platen, Ma, Jernite, Plu, Xu,
        Scao, Gugger, Drame, Lhoest, and Rush}]{Wolf2020}
  Thomas Wolf, Lysandre Debut, Victor Sanh, Julien Chaumond, Clement Delangue,
  Anthony Moi, Pierric Cistac, Tim Rault, Rémi Louf, Morgan Funtowicz, Joe
  Davison, Sam Shleifer, Patrick von Platen, Clara Ma, Yacine Jernite, Julien
  Plu, Canwen Xu, Teven~Le Scao, Sylvain Gugger, Mariama Drame, Quentin Lhoest,
  and Alexander~M. Rush. 2020.
  \newblock \href {https://www.aclweb.org/anthology/2020.emnlp-demos.6}
  {Transformers: State-of-the-art natural language processing}.
  \newblock In \emph{Proc. 2020 Conf. on Empirical Methods in Natural Language
    Processing: System Demonstrations}, pages 38--45, Online. Association for
  Computational Linguistics.

\end{thebibliography}
\end{document}